\documentclass[twoside,11pt]{article}

\usepackage{blindtext}
\usepackage{graphicx}%
\usepackage{multirow}%
\usepackage{amsmath,amssymb,amsfonts}%
\usepackage{amsthm}%
\usepackage{mathrsfs}%
\usepackage[title]{appendix}%

\usepackage{textcomp}%
\usepackage{manyfoot}%
\usepackage{booktabs}%
\usepackage{algorithm}%
\usepackage{listings}%
\usepackage{cite}
\usepackage{url}
\usepackage{adjustbox}
\usepackage{booktabs}
\usepackage{booktabs}
\usepackage{adjustbox,makecell}
\usepackage{amsmath,amsfonts}
\usepackage{array}
\usepackage{textcomp}
\usepackage{stfloats}
\usepackage{url}
\usepackage{verbatim}
\usepackage{graphicx}
\usepackage{amsmath}
\usepackage{amssymb}
\usepackage{booktabs}
\usepackage{adjustbox}
\usepackage{booktabs}
\usepackage{siunitx}
\usepackage{amsmath}
\usepackage[bb=px]{mathalfa} 
\usepackage{mathabx}
\usepackage{multirow}
\usepackage{siunitx}
\usepackage{soul}
\usepackage[accsupp]{axessibility}
\usepackage[absolute,overlay]{textpos}
\usepackage{subcaption}
\usepackage{amsmath}
\usepackage[table,xcdraw]{xcolor}
\usepackage[font=small]{caption}
\usepackage{amsmath,amssymb}
\usepackage{algorithm}
\usepackage{algpseudocode}
\usepackage{amssymb} 
\usepackage{tabularx}
\usepackage{graphicx}  
\usepackage{booktabs}  
\usepackage{xspace}
\usepackage{wrapfig}     
\usepackage{booktabs}

\usepackage{array}
\usepackage[inline]{enumitem}
\usepackage[abbrvbib, preprint]{jmlr2e}

\usepackage{lastpage}
\jmlrheading{23}{2022}{1-\pageref{LastPage}}{1/21; Revised 5/22}{9/22}{21-0000}{ Tajamul Ashraf and Iqra Altaf Gillani}

\ShortHeadings{Generalizable Federated Learning}{Ashraf et.al}
\firstpageno{1}

\begin{document}

\title{Generalizable Federated Learning using Client Adaptive Focal Modulation}

\author{\name Tajamul Ashraf \email tajamul.ashraf@mbzuai.ac.ae \\
       \addr Department of Computer Vision\\
       MBZUAI\\
       Abu Dhabi, Masdar City, UAE
       \AND
       \name Iqra Altaf Gillani \email iqraaltaf@nitsri.ac.in \\
       \addr Gaash lab, Department of Information Technology\\
       NIT Srinagar\\
       Srinagar, Kashmir, India}

\editor{My editor}

\maketitle

\def\noniid{\texttt{Non-IID}\xspace}
\def\fl{\texttt{FL}\xspace}
\def\pfl{\texttt{pFL}\xspace}
\def\cnns{\texttt{CNNs}\xspace}
\def\fedavg{\texttt{FedAvg}\xspace}
\def\localt{\texttt{Local-T}\xspace}
\def\fedavgt{\texttt{FedAvg-T}\xspace}
\def\focalnet{\texttt{FocalNet}\xspace}
\def\vit{\texttt{VIT}\xspace}
\def\mlp{\texttt{MLP}\xspace}
\def\model{\texttt{TransFED}\xspace}
\def\newmodel{\texttt{AdaptFED}\xspace}
\def\fsfda{\texttt{Fed-SFDA}\xspace}
\def\sfda{\texttt{SFDA}\xspace}
\def\sota{\texttt{SOTA}\xspace}
\def\da{\texttt{DA}\xspace}
\def\uda{\texttt{UDA}\xspace}
\def\fda{\texttt{FDA}\xspace}
\def\knn{\texttt{KNN}\xspace}
\def\cnn{\texttt{CNN}\xspace}
\def\cnns{\texttt{CNNs}\xspace}
\def\cfl{\texttt{CFL}\xspace}
\def\fedla{\texttt{pFedLA}\xspace}
\def\fedrod{\texttt{FedROD}\xspace}
\def\fedbn{\texttt{FedBN}\xspace}
\def\fedhn{\texttt{pFedHN}\xspace}
\def\fedgp{\texttt{pFedGP}\xspace}
\def\fedhnt{\texttt{pFedHN-T}\xspace}
\def\fedprox{\texttt{FedPROX}\xspace}
\def\fedper{\texttt{FedPER}\xspace}
\def\fedpert{\texttt{FedPER-T}\xspace}
\def\fedme{\texttt{pFedME}\xspace}
\def\ditto{\texttt{DITTO}\xspace}
\def\fedalign{\texttt{FedALIGN}\xspace}
\def\knnper{\texttt{KNNPer}\xspace}
\def\fedmd{\texttt{FedMD}\xspace}
\def\pfa{\texttt{PFA}\xspace}
\def\fedamp{\texttt{FedAMP}\xspace}
\def\fedtp{\texttt{FedTP}\xspace}
\def\fedtpt{\texttt{FedTP-T}\xspace}
\def\fedgen{\texttt{FedGEN}\xspace}
\def\dino{\texttt{DINO}\xspace}
\def\sgd{\texttt{SGD}\xspace}
\def\kd{\texttt{KD}\xspace}
\def\swa{\texttt{SWA}\xspace}

\begin{abstract}
Federated learning (\fl) has proven essential for privacy-preserving, collaborative training across distributed clients. Our prior work, TransFed, introduced a robust transformer-based FL framework that leverages a learn-to-adapt hypernetwork to generate personalized focal modulation layers per client, outperforming traditional methods in non-IID and cross-domain settings. In this extended version, we propose \newmodel, where we deepen the investigation of focal modulation in generalizable \fl by incorporating: (1) a refined adaptation strategy that integrates task-aware client embeddings to personalize modulation dynamics further, (2) enhanced theoretical bounds on adaptation performance, and (3) broader empirical validation across additional modalities, including time-series and multilingual data.
We also introduce an efficient variant of TransFed that reduces server-client communication overhead via low-rank hypernetwork conditioning, enabling scalable deployment in resource-constrained environments. Extensive experiments on eight diverse datasets reaffirm the superiority of our method over state-of-the-art baselines, particularly in source-free and cross-task federated setups. Our findings not only extend the capabilities of focal modulation in FL but also pave the way for more adaptive, scalable, and generalizable transformer-based federated systems. The code is avaliable at \href{http://github.com/Tajamul21/TransFed}{http://github.com/Tajamul21/TransFed}

\end{abstract}

\section{Introduction}
\label{1}
\textbf{Federated Learning.}
Federated Learning (\fl) is a prominent collaborative machine learning approach designed to utilize data from multiple clients without compromising their privacy~\cite{kaissis2020secure}, \cite{hsu2020federated}, \cite{zhang2021adaptive}, \cite{tran2019federated}. \fl is a widely used iterative learning method that maximizes the value of the data acquired across multiple clients~\cite{mcmahan2017communication}, \cite{ li2023edge}, \cite{li2023revisiting}.

However, traditional \fl methods, which aim to develop a single global model, often underperform due to the diverse data distributions and varying needs of different clients~\cite{li2020federated}, \cite{zhang2021edge}. In real-world scenarios, where datasets are \noniid (not independent and identically distributed), the significant differences across data can cause the global model to deviate substantially from local data patterns~\cite{hsieh2020non}. This mismatch can lead to slow convergence and poor inference performance, as the global model struggles to accommodate the unique characteristics of each client's data. Consequently, training a single global model proves to be ineffective when dealing with heterogeneous data and system variations across different clients~\cite{zhao2018federated}.

To deal with this challenge, personalized federated learning (\pfl)
mechanisms have been introduced that enable each client to train a model customized to their unique data distribution~\cite{fallah2020personalized}, \cite{hanzely2020lower}, \cite{huang2021personalized}, \cite{li2021ditto}. Recently, \pfl has gained significant attention for its ability to effectively manage statistical heterogeneity and enhance personalization in \fl~\cite{tan2022towards}, \cite{t2020personalized}, \cite{zhang2023fedcp}, 
\cite{zhang2023fedala}. This approach represents an evolution of traditional federated learning, focusing on developing individualized models for each client. 
\begin{wrapfigure}{r}{0.48\textwidth} 
    \centering
    \vspace{-10pt} 
    \includegraphics[width=0.48\textwidth]{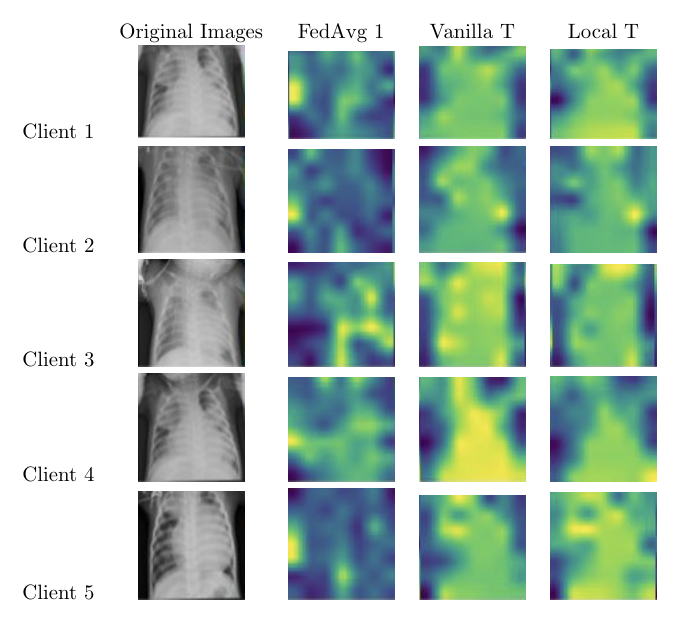}
    \caption{Comparing focal maps of \localt, \fedavgt, and Vanilla-T across clients, we see local training and Vanilla-T emphasize task details, while \fedavgt disrupts such information.}
    \label{img1}
    \vspace{-15pt} 
\end{wrapfigure}
By moving away from the single global model strategy, \pfl maintains the advantages of collaborative training while better accommodating the diverse needs of different clients~\cite{sattler2019robust},\cite{ sattler2020clustered}. An ideal \pfl system aims to achieve two main goals: aggregating information for collaborative learning and training reasonable personalized models. However, existing \pfl methods often prioritize one goal over the other, leading to imbalances. Strategies such as data-based approaches, single-model techniques like regularization~\cite{li2021ditto},\cite{ t2020personalized} and meta-learning~\cite{fallah2020personalized}, and multiple-model approaches like decoupling model parameters~\cite{collins2021exploiting},\cite{li2021fedbn},\cite{ liang2020think} have been proposed in the literature to address these challenges. Applying distance metrics to overall model parameters or loss values may not fully exploit client heterogeneity, hindering further personalization and impacting \pfl training efficiency. However, some \pfl frameworks heavily rely on Convolutional Neural Networks (\cnns), which can be sensitive to variations in diverse data~\cite{geirhos2018imagenet}. On the other hand, transformer models~\cite{vaswani2017attention} have shown exceptional performances in language and vision tasks.  Transformers utilize a self-attention mechanism that enables them to capture global interactions among inputs. This self-attention mechanism makes transformers more resilient to distribution shifts and data heterogeneity~\cite{ramachandran2019stand}.
Recognizing the varying utilities of different transformers is crucial for improving the effectiveness of \pfl methods.

\noindent\textbf{Transformers in Federated Learning.}
Inspired by the success of self-attention mechanisms, recent research has investigated employing transformers~\cite{vaswani2017attention} as the core network architecture for federated learning, in conjunction with the foundational federated averaging (\fedavg) algorithm~\cite{mcmahan2017communication}. While transformers have become the standard architecture in centralized computer vision tasks, their exploration in federated learning (\fl) remains limited. Qu et al.~\cite{qu2022rethinking} demonstrated the robustness of vision transformers against data heterogeneity in global \fl settings. In contrast, Sun et al.~\cite{sun2023fedperfix} and Li et al.~\cite{li2023fedtp} explored their personalization in \pfl. Further research has also focused on pre-trained transformers~\cite{chen2022importance},\cite{ nguyen2022begin}, \cite{sun2023fedperfix}. More recently, focal modulation networks~\cite{yang2022focal}, which replace self-attention with a new focal modulation technique to model input-dependent long-range interactions, have shown promising initial results. However, the impact of \fl algorithms on focal modulation requires further investigation, as it may limit the effectiveness of FocalNet-based transformers in downstream tasks. Recent studies have underscored the pivotal role of focal modulated layers in transformers~\cite{yang2022focal},\cite{ wang2023internimage},\cite{ zou2023generalized}, highlighting their superiority over other architectures. Given the potential of focal-modulated federated learning, we believe it is crucial to explore the challenges in this area comprehensively.  

\noindent\textbf{Main Challenges in Federated Learning.}
Building on this insight, we investigated focal modulation's impact on federated learning first. Our experiments included: 1) \localt, training individual Vision Transformer (\vit) models on each client; 2) \fedavg Transformer (\fedavgt), using \fedavg to train a global \focalnet model; and 3) Vanilla Tailored-T, maintaining local focal modulation with for server-side parameter aggregation.
We extend our prior investigation by conducting experiments on the \textbf{RSNA} pneumonia dataset~\cite{wang2017chestx}, a widely used medical imaging benchmark comprising chest X-rays labeled for pneumonia. The data was partitioned across five clients to simulate heterogeneous clinical environments and client-specific variation. However, as the number of clients increases, so does the communication burden between the clients and the central server, revealing a critical bottleneck in traditional federated settings.
To better understand model behavior, we generated attention maps using the Attention Rollout method~\cite{abnar2020quantifying}. As shown in Figure~\ref{img1}, both \localt and Vanilla Tailored-T models effectively highlighted diagnostically relevant regions (yellow heatmaps), whereas the baseline \fedavgt model failed to yield meaningful focal modulation attention~\cite{yang2022focal}. This underscores the limitations of directly applying focal modulation in federated environments.

Moreover, our extended analysis confirms that focal modulation layers lack scalability, as their number grows linearly with client count. Tailoring such layers for new clients requires complete retraining, which is computationally expensive and impractical for real-world deployment.
To address these limitations, \newmodel builds upon our earlier work TransFed~\cite{ashraf2024transfed}, a federated transformer framework that introduced a learn-to-tailor strategy for scalable client personalization. Rather than relying on conventional \focalnet-style local adaptation, Transfed utilizes a centralized learnable generator that dynamically produces client-specific focal modulation. In this extended version, we further explore and enhance the scalability and generalization capabilities of this approach across challenging cross-domain and source-free \fl settings, demonstrating its effectiveness in overcoming key limitations of focal modulation in federated environments.

\noindent\textbf{Domain Adaptation in Federated Setting.}
In federated learning, assuming labeled data on client devices is impractical due to high costs and manual effort~\cite{yao2022federated}. To address this, we focus on a practical scenario called Federated Source-Free Domain Adaptation (\fsfda). Here, the model is pre-trained using the server on labeled source data but cannot access it later, following the Source-Free Domain Adaptation (\sfda) setting~\cite{liu2021source}. Clients can only access their unlabeled target datasets, which they cannot share, reflecting real-world conditions with limited client data. After pre-training, the process transitions to a fully unsupervised approach. \fsfda aims to handle multi-target domain adaptation while addressing federated learning challenges like heterogeneity, communication bottlenecks, and client privacy. To our knowledge, previous works have not concurrently addressed this problem and its related issues. Our approach, \model, introduces a learnable generator on the server for matrix projections in focal modulation layers. These matrices enable adaptive queries, keys, and values, as well as parameter aggregation and sharing. By using the learn-to-tailor mechanism, \model efficiently distributes parameters and creates customized focal modulation layers. It achieves high accuracy, scalability with client numbers, and strong adaptability to new clients, making it a robust solution for \fsfda scenarios. The primary contributions of this work are summarized as follows:
\begin{itemize}
    \item We propose a refined \emph{task-aware adaptation strategy} that encodes client-specific context into the hypernetwork, enabling more fine-grained personalization and improved generalization across diverse domains.

\item To further push the boundaries of personalization in FL, we introduce \newmodel a lightweight, communication-efficient variant of \model, which leverages low-rank conditioning in the hypernetwork to scale across clients with limited bandwidth.

\item We expand the evaluation of \model to a broader range of tasks, including \emph{multilingual language modeling}, \emph{time-series forecasting}, and \emph{multi-modal vision-language tasks}, demonstrating consistent improvements under severe domain shift and limited supervision.

\item We revisit and extend the \fsfda setup, Federated Source-Free Domain Adaptation, by incorporating a learnable distribution-aware generator that handles unlabeled, non-iid client data in cross-domain FL setups without access to source data.

\item Our theoretical analysis offers tighter adaptation bounds for the learn-to-adapt strategy, shedding light on how hypernetwork-driven modulation achieves robust personalization in non-IID federated learning.

\end{itemize}


\section{Related Work}
\label{2}

\noindent\textbf{Personalized Federated Learning (\pfl).}
In recent years, numerous approaches have been developed to tackle heterogeneity in personalized federated learning (\pfl). These methods generally fall into two broad categories: data-based and model-based strategies. Data-based approaches aim to mitigate statistical heterogeneity by modifying client datasets, while model-based methods focus on adapting model architectures or parameters to suit individual clients \cite{zhang2022r, zhang2022federated}.  
To address challenges related to data heterogeneity and privacy in \pfl, researchers have proposed a range of techniques \cite{wang2019federated, mansour2020three, fallah2020personalized}. Data-based strategies often involve augmenting local datasets with a small shared global dataset or implementing client selection mechanisms to ensure a more homogeneous data distribution \cite{zhang2022r, zhang2022semi, zhang2022federated}. Model-based approaches can be further classified into single-model and multiple-model frameworks.  

Single-model methods extend conventional federated learning (\fl) techniques by incorporating local fine-tuning, regularization, meta-learning, model mixtures, or parameter decomposition \cite{wang2019federated, mansour2020three, fallah2020personalized}. In contrast, multiple-model approaches train several global models tailored for heterogeneous clients or facilitate collaboration to personalize models at an individual level \cite{li2020federated, t2020personalized, li2021ditto}.  Several techniques enhance personalization by fine-tuning global models on local client datasets, introducing proximal regularization, leveraging knowledge distillation, or using clustering methods to group similar clients \cite{mendieta2022local, zhu2021data, sattler2020clustered, dosovitskiy2020image, arivazhagan2019federated}. Fine-tuning the global model for each client is a common strategy for obtaining personalized parameters \cite{wang2019federated, mansour2020three, fallah2020personalized}. Proximal regularization, as implemented in methods like \fedprox \cite{li2020federated}, \fedme \cite{t2020personalized}, and \ditto \cite{li2021ditto}, helps mitigate client drift by constraining updates during training. \fedalign \cite{mendieta2022local} tackles data heterogeneity from a generalization perspective, ensuring models remain robust across diverse client distributions.  

Hybrid approaches have also emerged, such as \knnper \cite{marfoq2022personalized}, which integrates k-nearest neighbors with existing \fl models to enhance personalization while preserving local feature representations. Knowledge distillation techniques like \fedmd \cite{li2019fedmd} and \fedgen \cite{zhu2021data} use global teacher models to guide the training of client-specific models. Clustered \fl methods, including \cfl \cite{sattler2020clustered}, \pfa \cite{liu2021pfa}, and \fedamp \cite{huang2021personalized}, leverage client similarities to train models more effectively within homogeneous subgroups.  
More recently, transformer-based approaches have been explored for federated personalization. \fedtp \cite{li2023fedtp} focuses on adapting attention maps in vision transformers (\vit) \cite{dosovitskiy2020image} by decoupling cross-attention mechanisms to manage data heterogeneity. Another method, FedREP \cite{arivazhagan2019federated}, improves personalization by training local classifier heads while maintaining shared base layers across clients. Overall, these advancements highlight the growing sophistication of federated learning in addressing client heterogeneity, paving the way for more adaptive and efficient decentralized learning systems.

\noindent\textbf{Federated-based Domain Adaptation}
Vision tasks often require dense annotations, but recent methods use synthetic data from virtual environments to reduce costs \cite{toldo2020unsupervised, richter2016playing, alberti2020idda, testolina2023selma}. However, models trained on synthetic data struggle to generalize due to domain shifts. Domain adaptation (\da) aims to bridge the gap between source and target domains, particularly in Unsupervised Domain Adaptation (\uda) scenarios where target data lacks labels. Early \da methods focused on reducing domain divergence \cite{long2015learning, saito2018maximum}, with adversarial training becoming popular involving networks and domain discriminators \cite{tsai2018learning, luo2019taking}. Non-trainable style translation algorithms like \fda~\cite{yang2020fda} have been introduced to address these challenges. Modern approaches \cite{lian2019constructing, zou2018unsupervised, hoyer2022daformer} use self-learning to generate pseudo-labels from target data, enabling fine-tuning in federated settings where clients only have access to unlabeled data.

Our proposed \model framework employs a learn-to-customize mechanism to train tailored focal-modulation layers within a transformer, effectively addressing client data heterogeneity. 
\begin{wrapfigure}{r}{0.5\linewidth} 
  \vspace{-8pt}
  \centering
  \includegraphics[width=\linewidth]{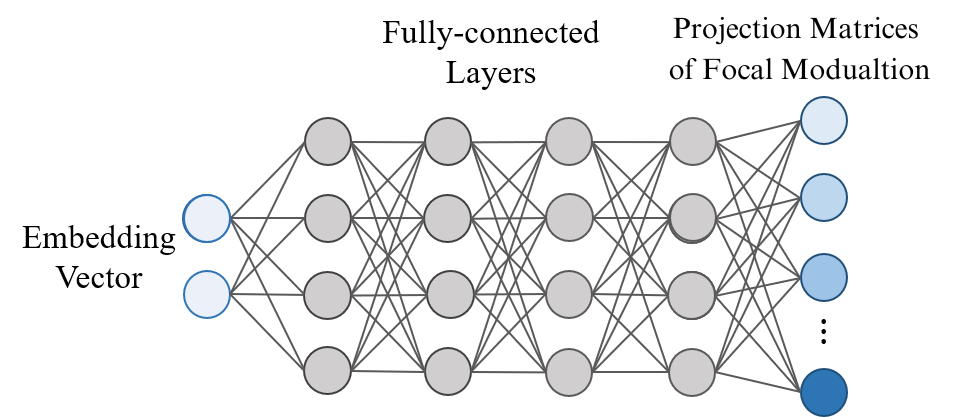}
  \caption{Structure of Hypernetwork}
  \label{hyper}
  \vspace{-10pt}
\end{wrapfigure}
Unlike traditional methods, \model uses a server-based trainable generator to produce projection parameters within focal-modulation layers, enabling client-specific
queries, keys, and values. 
This approach builds on transformer advancements in vision tasks \cite{dosovitskiy2020image} and leverages findings that transformers perform better than \cnns in heterogeneous federated learning \cite{qu2022rethinking}. \model distinguishes itself by integrating a learnable generator within a transformer, contrasting with \cnn-focused methods like \fedhn~\cite{ha2016hypernetworks}, \fedla~\cite{ma2022layer}, and \fedrod~\cite{chen2021bridging}.
The study of Domain Adaptation (\da) in Federated Learning (\fl), including Unsupervised Domain Adaptation (\uda) and especially Source-Free Domain Adaptation (\sfda), is still emerging. Additionally, Clustered FL (\cfl) approaches group clients with similar characteristics, such as geographic location, to create personalized models for specific conditions~\cite{kundu2021generalize}, \cite{fallah2020personalized}, \cite{ghosh2020efficient}. Unlike these methods, we use the adaptive learnable generator to bridge cross-domain shifts.

\section{Federated Learning by Tailored Focal Modulation}
\label{3}

This section introduces \newmodel, which is specifically designed to address heterogeneity in a source-free setting and produces personalized, high-quality adaptive models for individual clients.

\subsection{Problem Statement}
\label{3.1}
In the context of visual tasks, the earlier \model~\cite{ashraf2024transfed} framework incorporates the utilization of a traditional \dino model~\cite{caron2021emerging}. The initial step in processing the input sequence $S$, which has a fixed length $l$, involves partitioning the images into a sequential format during the image preprocessing phase of the \focalnet. Subsequently, this sequential representation is converted into an embedding matrix $M$, with dimensions $\mathbb{R}^{n \times m}$. The focal-modulation mechanism operates on the queries, keys, and values, denoted as $Q = MP^Q$, $K = MP^K$, and $V = MP^V$, respectively. We concatenate these projection parameters into $P = [P^Q, P^K, P^V]$ for simplicity.

By utilizing a visual feature map $X \in \mathbb{R}^{H \times P \times C}$ as the input, a standard encoding process produces a feature representation $y_i \in \mathbb{R}^C$ for each visual token (query) $Q_i \in \mathbb{R}^C$. This generation is accomplished through the token's interaction with its surroundings, including neighboring tokens, and the aggregation of information across contexts. The process involves the interaction function $\tau$ and the aggregation function $A$. Consequently, the refined representation $y_i$ is obtained by combining the aggregated context features, obtained through the function $A$ at each location $i$, with the query $Q_i$ through the interaction function $\tau$.
Focal Modulation generates refined representation \( y_i \) using an early aggregation procedure formulated as:
\begin{figure*}[t]
    \centering
    \includegraphics[width= \textwidth]{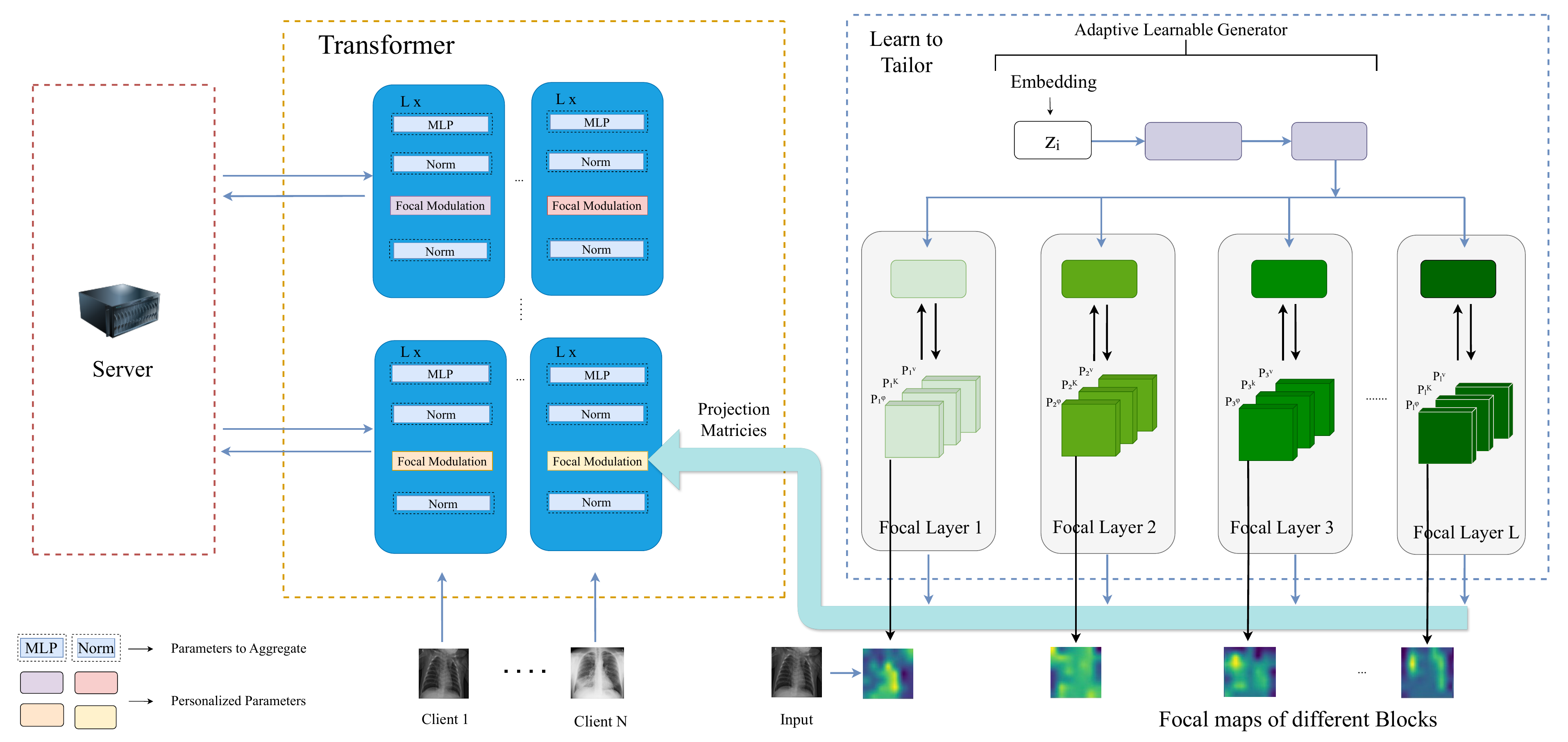}
    \caption{\newmodel Overview: The architecture combines focal modulation layers and centralized parameter aggregation to enhance collaborative learning while preserving personalization. Clients retain self-attention layers locally, while a server-based hypernetwork generates personalized projection matrices, ensuring adaptation to diverse data distributions.}
    \label{model}
\end{figure*}

\begin{equation}
\small
y_i = \tau(A(i, X), Q_i) 
\end{equation}
where the context features are first aggregated using $A$ at each location $i$, then the query interacts with the aggregated feature based on $\tau$ to form $y_i$.
To emulate a federated scenario, we examine a collection of $N$ clients designated by $[N]$, with each specific client $i$ holding its own local dataset$D_i = {(x_{i}^{(j)}, {y_{i}^{(j)})}}_{j=1}^{m_i}$ ($1 \leq i \leq N$), consisting of $m_i$ samples drawn from a distinct data distribution $P_i$. The total dataset is denoted as $D = \bigcup_{i \in [N]} D_i$, with a total size of $S = \sum_{i=1}^{N} m_i$.
The customized model associated with client $i$, defined by the parameters $\theta_i$, is denoted as $f(\theta_i; \cdot)$. The optimization objective is defined as follows:

\begin{equation}
\small
\arg \min \limits_{\substack{\Theta}} \sum_{i=1}^{N} \limits \left(\frac{m_i}{S}\right) K_i(\theta_i)
\end{equation}

where $K_i(\theta_i) = \mathbb{E}_{(x{(j)i},y_{(j)i})_\in D_i} [l(f(\theta_i;(x_{i}^{(j)}), (y_{i}^{(j)})]$.
Here,$\Theta$ = ${\theta_i}_{i=1}^N$ represents the set of tailored parameters for each client, and $l(\cdot, \cdot)$ denotes the per-sample loss function that is common to all clients. The selection of the loss function for a specific task, whether it is a mean square error or cross-entropy loss, is contingent upon the nature of the task.

\begin{algorithm}[t]
\caption{TransFed: Generalizable Focal Modulation}
\label{algo}
\small
\begin{algorithmic}[1]
\Require $C$ \text{ (rounds)}, $L$ \text{ (local epochs)}, $\alpha$ \text{ (local LR)}, $\beta$ \text{ (global LR)}.
\State Initialize parameters $\xi^{0},\, z_i^{0}$ for all clients $i$, and $\phi^{0}$
\For{$c = 1$ to $C$}
    \State Sample client set $C_c \subset \{1,\ldots,N\}$
    \For{$i \in C_c$}
        \State $\xi_i^{\,c,0} \gets \bar{\xi}^{\,c-1}$ \Comment{broadcasted init for client $i$}
        \State $P_i^{\,c,0} \gets h(\phi^{\,c-1}, z_i^{\,c-1})$
        \State $\theta_i^{\,c,0} \gets \{P_i^{\,c,0},\, \xi_i^{\,c,0}\}$
        \For{$k = 0$ to $L-1$}
            \State Sample mini-batch $B_i \subset D_i$
            \State $\theta_i^{\,c,k+1} \gets \theta_i^{\,c,k} - \alpha \,\nabla_{\theta_i}\,\mathcal{L}_i\!\big(\theta_i^{\,c,k}; B_i\big)$
        \EndFor
        \State $\Delta P_i \gets P_i^{\,c,L} - P_i^{\,c,0}$
    \EndFor
    \State $\displaystyle \bar{\xi}^{\,c} \gets \sum_{i \in C_c} \frac{m_i}{\sum_{j \in C_c} m_j}\; \xi_i^{\,c,L}$
    \State $\displaystyle \phi^{\,c} \gets \phi^{\,c-1} \;-\; \beta \sum_{i \in C_c} \frac{m_i}{\sum_{j \in C_c} m_j}\; \nabla_{\phi}\, P_i^{\,c}\, \Delta P_i$
    \Comment{global meta-params}
    \For{$i \in C_c$}
        \State $\displaystyle z_i^{\,c} \gets z_i^{\,c-1} \;-\; \beta \,\nabla_{z_i}\, P_i^{\,c}\, \Delta P_i$
        \Comment{client-specific aux update (no sum over $i$)}
    \EndFor
\EndFor
\State \Return $\bar{\xi}^{\,C},\; \phi^{\,C},\; \{z_i^{\,C}\}_{i=1}^N$
\end{algorithmic}
\end{algorithm}

\subsection{Vanilla Tailoring of Focal Modulation}
\label{3.2}
Federated learning's popularity stems from global insights via focal modulation layers. However using \newmodel on client layers can harm performance with diverse data. To address this challenge, our solution involves tailored focal modulation. This approach entails customizing certain local layers while averaging other layers to maintain standard insights.

In \newmodel, parameters are locally trained and aggregated on the server, similar to \fedavg. The focal-modulation layer has projection parameters $P_i$, while other layers have parameters $\xi$. The tailored model, denoted as $\theta_i = (P_i, \xi)$, undergoes local training. This process is repeated for multiple communication rounds. Resulting in the updated model $f(P_{i}^{t,k}, \bar{\xi}_{i}^{t,k}; \cdot)$, where $P_{i}^{t,k}$ is retained locally to store the tailored information of client $i$, and $\bar{\xi}_{i}^{t,k}$ is aggregated across the clients using Equation (2):

\begin{equation}
\small
\bar{\xi}^{t} = \sum_{i=1}^{N} \limits  \left(\frac{m_i}{S}\right)  \xi^{t,k}_i
\end{equation}

Consequently, the objective function of \newmodel, as derived from Equation (1), is to minimize the following loss:

\begin{equation}
\small
\arg \min \limits_{\substack{\theta}} \sum_{i=1}^{N} \limits \left(\frac{m_i}{S}\right) K_i(P_i, \xi)
\end{equation}

where

\begin{equation}
\small
K_i(P_i, \xi) = \sum_{i=1}^{N} \limits  \left(\frac{m_i}{M}\right) \mathbb{E}_{({x_i^j, y_i^j}) \in D_i} l(f(P_i, \xi; x_i^{(j)}), y_i^{(j)})
\end{equation}

While the vanilla customization procedure generates tailored focal-modulation layers through local training, it overlooks the potential inherent client relationships, leading to sub-optimal tailored models. Moreover, the scalability of tailored focal-modulation layers becomes an issue as the number of clients grows linearly. Additionally, the adaptation capability of tailored Focal Modulation is limited, requiring retraining when novel clients are introduced to obtain specific focal modulation layers for them.
\begin{wrapfigure}{r}{0.6\columnwidth}
    \centering
    \includegraphics[width=0.58\columnwidth]{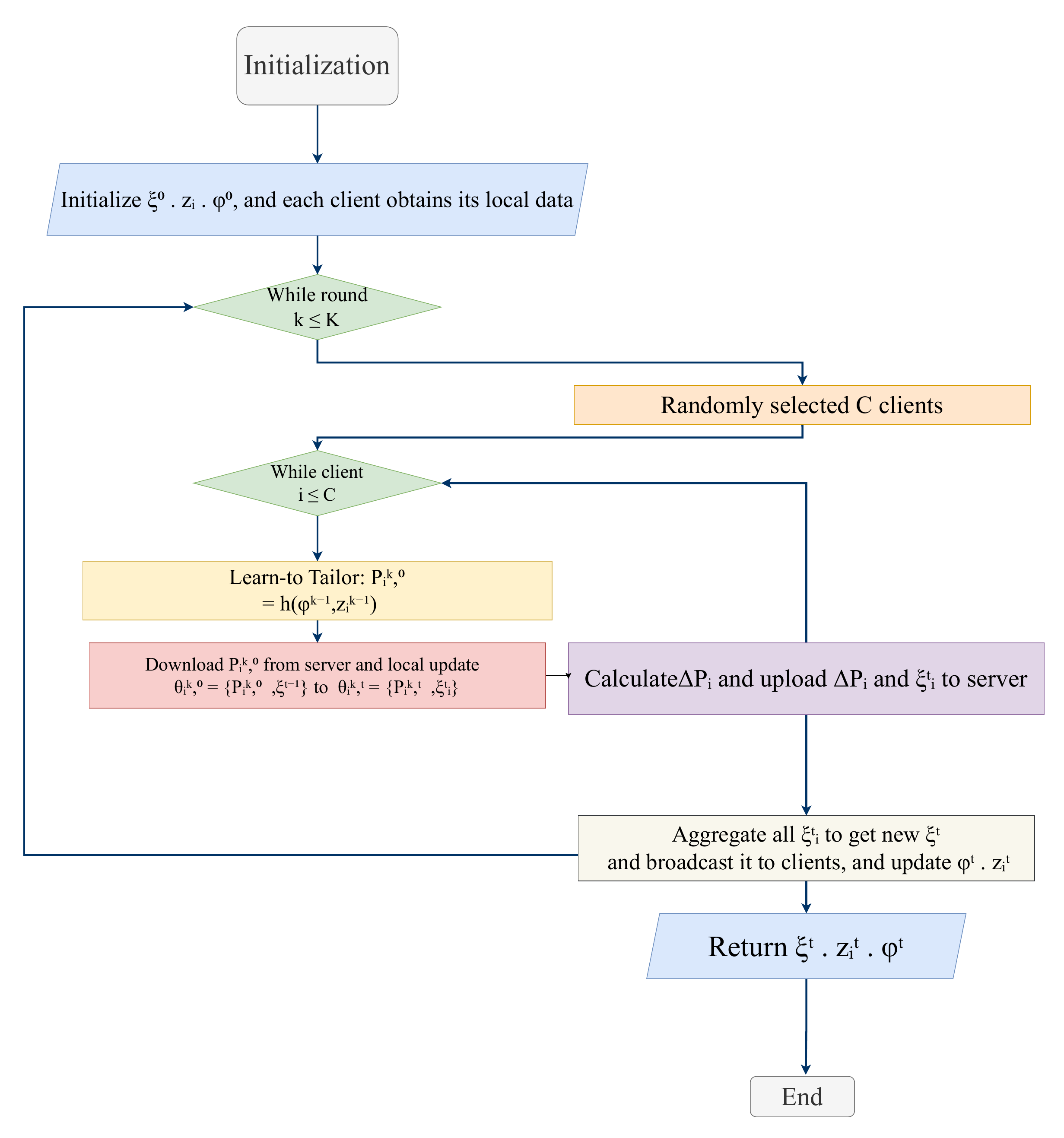}
    \caption{Visualization of Client Embeddings Learned by \newmodel using t-SNE on the RSNA Dataset.}
    \label{fig:tsne2}
      \vspace{-39pt} 
\end{wrapfigure}
\subsection{Custom Learning for Focal Modulation}
\label{3.3}
This section introduces \newmodel, a framework that incorporates a learn-to-tailor approach to augment the existing vanilla customization mechanism for focal Modulation.

\noindent Algorithm \ref{algo} presents the \newmodel process for parameter updates in a federated learning scenario. It spans communication rounds ($C$) and local epochs ($L$), iterating through clients to locally update model parameters ($\theta$) using mini-batches. Global parameters $\phi$ and $z_i$ are also updated collectively, yielding refined global parameters $\bar{\xi}^t$, $\phi^t$, and $z^t$. These enhancements encourage effective collaboration among clients while retaining individual data characteristics. In the \newmodel approach, an adaptive learnable generator \cite{ha2016hypernetworks} is integrated into the server's functionality, generating projection matrices intended for the focal-modulation layers of individual clients (as illustrated on the right side of Figure \ref{model}). This design facilitates the effective sharing of parameters among the clients.

The adaptive learnable generator at the server, denoted as $h(\phi; z_i)$ and parameterized by $\phi$, takes as input an adaptive learnable embedding vector $z_i \in \mathbb{R}^D$ associated with client $i$, which can either be a client-specific embedding or a fixed vector. We implement the adaptive learnable generator using simple fully connected layers, where each transformer block's last layer is unique. The $z_i$ vector, the adaptive learnable generator produces the projection parameters $P_i = h(\phi; z_i)$ for client $i$, which are decomposed into the query, key, and value projection matrices for the focal-modulation mechanism, denoted as $P_i = [P_{Qi}, P_{Ki}, P_{Vi}]$.

This approach enables the adaptive learnable generator to learn a set of projection parameters ${P_i = h(\phi; z_i) | 1 \leq i \leq N}$ for tailored Focal Modulation. Consequently, the tailored model is represented as $f(P_i, \xi; \cdot) = f(h(z_i; \phi), \xi; \cdot)$, and loss function is updated as follows:
\begin{equation}
K_i(P_i, \xi) = K_i(h(\phi; z_i), \xi) 
= \sum_{i=1}^{N} \left(\frac{m_i}{M}\right) 
  \mathbb{E}_{(x_i^j, y_i^j) \in D_i} 
  l\big(f(h(\phi; z_i), \xi; x_i^{(j)}), y_i^{(j)}\big)
\end{equation}

This updated loss function computes the training loss for client $i$ by applying the adaptive learnable generator-generated projection parameters $P_i = h(\phi; z_i)$ alongside the standard parameters $\xi$ to the tailored model.
The update mechanism within each epoch is represented by the variable $k$, and the local model parameter $\theta_i$ is subjected to updates through stochastic gradient descent (\sgd), as defined by the following equation:
\begin{equation}
\small
\theta_i^k \leftarrow \theta_i^{k-1} - \alpha \nabla_{\theta_i} K_i(\theta_i^{k-1}; B_i)
\end{equation}
where $B_i$ represents a mini-batch extracted from $D_i$.

Representing the collection of selected clients in each round $t$ as $C_t$. The gradients of $\phi$ and $z_i$ can be obtained from Equation (6) using the chain rule:
\begin{equation}
\small
\nabla_\phi K_i
  = \sum_{i \in C^t} \left(\frac{m_i}{M}\right)
    \nabla_\phi P_i^{\top} \, \Delta_{p_i}
\end{equation}

\begin{equation}
\small
\nabla_{z_i} K_i
  = \sum_{i \in C^t} \left(\frac{m_i}{M}\right)
    \nabla_{z_i} p_i^{\top} \, \Delta_{P_i}
\end{equation}

where $\triangle P_i = P_{Ki} - P_{Qi}$ represents the change in projection parameters after $K$ epochs of local updates.

During communication round $t$, updates are applied to the adaptive learnable generator parameter $\phi$ and the client embedding $z_i$ through the utilization of computed gradients:
\begin{equation}
\phi^{t} = \phi^{t-1} - \beta \nabla_{\phi} K_i^{(t-1)}
\end{equation}
\begin{equation}
z_i^{t} = z_i^{t-1} - \beta \nabla_{z_i} K_i^{(t-1)}
\end{equation}

Contrasted with the standard customization method, the learn-to-tailor approach within \newmodel brings forth a range of benefits. Firstly, it achieves effective parameter sharing across clients while harnessing the focal modulation mechanism's potency in federated learning. Secondly, its scalability accompanies an expanding client base, owing to the shared adaptive learnable generator with personalized embedding vectors driving the focal modulation layer's generation. Lastly, its adaptability extends to new clients whose data remains unseen during training. The initial and final aspects will undergo validation in Section IV. The middle aspect is substantiated by a comparison of parameter counts between learn-to-customize and standard customization. The adaptive learnable generator, adopting an MLP architecture, comprises parameters roughly equivalent to $D_h \times D_s$, where $D_h$ and $D_s$ denote the dimensions of the hidden layers within the adaptive learnable generator and the self-attention projection parameters, respectively. In contrast, the cumulative self-attention projection parameters in the standard customization rise linearly with the client count, i.e., $N \times D_s$. With a substantial client count ($N > D_h$), learn-to-customize for focal modulation consumes fewer resources.

\subsection{\fsfda Setting}

In this section, we formalize the proposed Federated Source-Free Domain Adaptation (\fsfda) setting. We consider a central server and a set of clients \( K \) with \( |K| = K \). The source dataset \( D_S \), kept on the server, consists of image-segmentation label pairs \((x_S, y_S) \sim P_S(x, y)\), where \( x_S \in X \) and \( y_S \in Y^{N_p} \). Each client \( k \in K \) has a target training dataset \( D_{T_k} \) with images \( x_{T_k,i} \sim P_{T_k}(x) \). The source and test datasets share the same categories \( Q = Q_S = Q_T \). Clients' local datasets, drawn from a meta-distribution with \( G \) latent visual domains, may have similar distributions. The test dataset \( D_{T_{\text{test}}} \) follows the target distribution \( P_T \) and evaluates the final model.

The first step of \fsfda involves pre-training on \( D_S \). To align source and target styles, we apply \fda \cite{yang2020fda} style transfer. Clients extract and send their style \( s_k \) to the server, which augments \( D_S \) with these styles for training. After pre-training, \( D_S \) is no longer used.
Pseudo-labels help mimic label presence but can lead to overconfidence and misclassifications. To address this, we use a Knowledge Distillation (\kd) loss \( \mathcal{L}_{\kd} \) to retain pre-trained knowledge. However, \kd alone can lead to slight overfitting. To mitigate this, we apply Stochastic Weight Averaging (\swa)~\cite{} to clients’ teachers \( g_c \).

\[
w_{t+\omega}^{g_c} = \frac{w_t^{g_c} n_t^{g_c} + w_{t+\omega}^{c}}{n_t^{g_c} + 1}
\]
where \( n_t^{g_c} = \frac{t - t_{\text{START}}}{\omega} \). This technique, called \swa teacher, reduces noise and stabilizes the learning curve, enabling better convergence to the local minimum of the total loss:
\[
L = L_{\texttt{PSEUDO}} + \lambda_{\text{\kd}} L_{\text{\kd}}
\]
where \( L_{\texttt{PSEUDO}} \) is the pseudo-label loss and \( \lambda_{\kd} \) is a hyperparameter controlling the \kd loss.

\subsection{Generalization Bound}
We analyze the generalization bound of \newmodel in this section. Before starting the analysis, we first introduce some assumptions as follows. 

\noindent \textbf{Assumption 1}
\emph{The embedding vectors $z_i$ and the weights $\varphi$ of the hypernetwork $h(\varphi, z_i)$ are assumed to lie within a bounded region with radius $R_h$. Similarly, the parameters $\xi$ of other layers in the Transformer are constrained within a bounded region of radius $R_t$. Formally, these constraints can be expressed as:
\begin{equation}
\Vert \varphi-\varphi^{\prime} \Vert \leq R_h,\ \Vert z_i-z_i^{\prime} \Vert \leq R_h,\ \Vert \xi-\xi^{\prime} \Vert \leq R_t.
\end{equation}
}

\noindent \textbf{Assumption 2}
\emph{(Lipschitz Conditions) Let $\mathcal{D}_1, \mathcal{D}_2, \dots, \mathcal{D}N$ denote the real data distributions. Define the expected loss as $\mathcal{L}{\mathcal{D}i}(h(\varphi; z_i), \xi) = \mathbb{E}{(x, y) \in \mathcal{D}_i} l(f(h(\varphi; z_i), \xi; x), y)$. We assume that the following Lipschitz conditions are satisfied:}
\begin{subequations}
\begin{equation}
|\mathcal{L}_{\mathcal{D}_i}(h(\varphi;z_i),\xi)-\mathcal{L}_{\mathcal{D}_i}(h(\varphi;z_i),\xi^{\prime})|  \leq L_{\xi} \Vert \xi-\xi^{\prime} \Vert,
\end{equation}
\begin{equation}
\begin{aligned}
&|\mathcal{L}_{\mathcal{D}_i}(h(\varphi;z_i),\xi)-\mathcal{L}_{\mathcal{D}_i}(h^{\prime}(\varphi;z_i),\xi)| \\
&\leq L_h \Vert h(\varphi;z_i)-h^{\prime}(\varphi;z_i)\Vert,
\end{aligned}
\end{equation}
\begin{equation}
\Vert h(\varphi;z_i)-h(\varphi^{\prime};z_i)\Vert \leq L_{\varphi} \Vert \varphi-\varphi^{\prime} \Vert,
\end{equation}
\begin{equation}
\Vert h(\varphi;z_i)-h(\varphi;z_i^{\prime})\Vert \leq L_z \Vert z_i-z_i^{\prime} \Vert.
\end{equation}
\end{subequations}

\noindent \textbf{Theorem 1}
\emph{Let ${\mathcal{\hat{D}}}_1, {\mathcal{\hat{D}}}_2, \dots, {\mathcal{\hat{D}}}_N$ represent the empirical data distributions for $N$ clients. The parameters $\hat{\varphi}$, $\hat{z}_i$, and $\hat{\xi}$ are obtained by training on these empirical distributions. Define $\mathcal{H}$ as the set of personalized hypotheses, and let $d$ denote the VC-dimension of $\mathcal{H}$. Assuming that Assumptions 1 and 2 are satisfied, the following holds with probability at least $1 - \delta$:}

\begin{equation}\label{Generalization Bound}
\resizebox{0.9\linewidth}{!}{$
\begin{aligned}
    &\left|\sum_{i=1}^{N}\frac{m_i}{M}\mathcal{L}_{\hat{\mathcal{D}}_i}(h(\hat{\varphi};\hat{z_i}),\hat{\xi}) - \sum_{i=1}^{N}\frac{m_i}{M}\mathcal{L}_{\mathcal{D}_i}(h(\varphi^*;z_i^*),\xi^*)\right| 
    &\leq \sqrt{\frac{M}{2}\log\frac{N}{\delta}} + \sqrt{\frac{dN}{M}\log\frac{eM}{d}} + L_h R_h (L_{\varphi}+L_z) + L_{\xi} R_t
\end{aligned}
$}
\end{equation}

\noindent\emph{where 
$\varphi^*$, $z_i^*$, and $\xi^*$ represent the optimal parameters corresponding to the real distribution of each client, respectively; the size of the whole dataset is $M$ with the local data size of client $i$ being $m_i$. 
}

Theorem 1 highlights that the performance of a model trained on empirical distributions is influenced by the complexity of the hypothesis class $\mathcal{H}$ (quantified by its VC-dimension), the number of clients, the total dataset size, and the associated Lipschitz constants.

{\color{black} The second term on the right-hand side of (\ref{Generalization Bound}) can be expressed as $\sqrt{\frac{\log (eM/d)}{M/dN}}$, indicating its dependency on the ratio $\frac{M}{d}$. We define the hypothesis classes of \newmodel with the \emph{learn-to-personalize} mechanism and vanilla personalization as $\mathcal{H}_h$ and $\mathcal{H}_v$, respectively. The VC-dimension of $\mathcal{H}_h$ is smaller than that of $\mathcal{H}_v$, particularly when the number of clients is large. This reduction arises because \newmodel employs a single hypernetwork to generate self-attention layers for all clients via the \emph{learn-to-personalize} mechanism as shown in Figure~\ref{hyper}. As the VC-dimension $d$ decreases, the value of $\sqrt{\frac{\log (eM/d)}{M/dN}}$ also reduces. Consequently, \newmodel achieves better generalization compared to vanilla personalization for self-attention layers, especially in scenarios with a large number of clients.}

The detailed key lemmas and the proof of Theorem 1 are provided in the Appendix below.
\section{Experiments}
\label{4}

This section introduces the experimental configuration, assesses our proposed model's performance, and compares several baseline methods across diverse learning scenarios. We introduce the benchmarks, \noniid settings, model architectures used in our experiments, and relevant implementation details.
The performance evaluation focuses on analyzing different aspects of our model, including the impact of network backbones and customized components of the transformer, the adaptation ability to novel clients, the compatibility of \newmodel with other methods, and the visualization of attention maps. Additionally, we investigate the effect of heterogeneity in label distribution, noise-induced feature imbalances, and the influence of different parameters.
To provide a clear overview of the experimental framework, Figure \ref{model} depicts the flowchart of our \newmodel approach. This section outlines our model's experimental design, performance evaluation metrics, and key findings, providing valuable insights into its effectiveness and versatility in various learning scenarios.

\subsection{Experimental Setup}

\begin{table}[h]
\centering
\caption{Datasets and Models.}
\label{tab_datasets}
\resizebox{\columnwidth}{!}{%
\begin{tabular}{l | c c c l}
\toprule
\rowcolor[gray]{0.90}
Dataset & Task & Clients & Samples & Model \\
\midrule
\textbf{RSNA} \cite{wang2017chestx} & Image. Classification. & 100/200 & 30,227 & FocalNet \\
\textbf{Kermany} \cite{kermany2018labeled} & Image. Classification. & 100/200 & 5,232 & FocalNet \\
\textbf{Shakespeare} \cite{shakespeare} & Character. Prediction. & 683 & 2,578,349 & LSTM, Transformer \\
\textbf{CIFAR-10} \cite{krizhevsky2009learning} & Image. Classification. & 100/200 & 60,000 & FocalNet \\
\textbf{CIFAR-100} \cite{krizhevsky2009learning} & Image. Classification. & 100/200 & 60,000 & FocalNet \\
\textbf{Office-Caltech} \cite{gong2012geodesic} & Image. Classification. & 50/100 & 2,533 & FocalNet \\
\bottomrule
\end{tabular}%
}
\end{table}

\subsubsection{Baselines}
Our evaluation comprehensively compared the \newmodel with various federated learning algorithms. We compared \newmodel against fundamental federated algorithms, including \fedavg \cite{mcmahan2017communication} and \fedprox  \cite{li2020federated}. Additionally, we evaluated its performance against state-of-the-art (\sota) customization algorithms, including \fedper \cite{arivazhagan2019federated}, \fedme \cite{t2020personalized}, and \fedtp \cite{li2023fedtp}, as well as Vanilla-based models. Also, we compare our method with state-of-the-art source-free domain adaptation (\sfda) methods~\cite{feng2023robust}. By including these various algorithms in our comparison, we aimed to assess the effectiveness and superiority of \newmodel in achieving customized and efficient federated learning. The selected algorithms represent a range of approaches that tackle different aspects of customization in federated learning, allowing us to evaluate \newmodel's performance with basic and advanced techniques.
This comprehensive evaluation provides valuable insights into the relative strengths and weaknesses of \newmodel compared to existing state-of-the-art algorithms, further highlighting its potential as an advanced customization approach in the domain of federated learning.

\subsubsection{Datasets}
We conducted experiments on five different diverse datasets as shown in Table~\ref{tab_datasets}. Experiments were conducted on two widely used pneumonia benchmark datasets: \textbf{Kermany} \cite{kermany2018labeled} and \textbf{RSNA} \cite{wang2017chestx}. Additionally, to validate our approach on other domains, we performed experiments on the publicly available datasets \textbf{CIFAR-10}~\cite{krizhevsky2009learning}, \textbf{CIFAR-100}~\cite{krizhevsky2009learning}, and \textbf{Shakespeare}~\cite{mcmahan2017communication}. In addition to this, we compare our method within the Source-Free Domain Adaptation (\sfda) setting on \textbf{Office-Caltech}~\cite{gong2012geodesic} for image classification tasks in a federated learning (\fl) setup.

We utilized two partitioning techniques to simulate non-identically distributed (\noniid) scenarios in our experiments. These techniques were applied to four image datasets and one language dataset.
For the image datasets, we employed two distinct split strategies to create \noniid conditions. The first strategy, known as the \emph{Pathological} setting, involved assigning classes to each client in a randomized manner. Specifically, for a given class $c$ and client $i$, the sample rate was determined by $a_{i,c}/\sum_j a_{j,c}$, where $a_{i,c}$ is a value drawn from a uniform distribution $U(0.4, 0.6)$. 
The second strategy involved creating a federated version of the datasets by partitioning samples with the same labels across clients. This was achieved using a symmetric \emph{Beta distribution} with a parameter of $\alpha=0.3$. This approach ensured that samples were distributed among clients in a federated manner, promoting a diverse distribution of data for training.
To enhance the realism of the local datasets within the \textbf{Kermany} dataset and \textbf{CIFAR-100}, we employed a two-stage Pachinko allocation method. Initially, a Beta distribution with a parameter of $\alpha=0.4$ was generated over coarse labels for each client. In the second stage, a Beta distribution with a parameter of $\beta=10$ was generated over the fine labels corresponding to the coarse labels. The class distribution and the allocation of classes in both the training and test sets were kept consistent for the coarse and fine label partitions among clients. Table \ref{tab_datasets} provides a summary of the datasets, their associated tasks, and the counts of clients and models involved.

For the \textbf{Shakespeare} dataset split in training and testing, we used the standard 80\%-20\% split, following established practices.
Additionally, we utilized the \textbf{Office-Caltech} dataset, which consists of 10 common classes shared by the \textbf{Office-31}~\cite{saenko2010adapting} and \textbf{Caltech-256}~\cite{griffin2007caltech} datasets. This dataset encompasses four domains: Amazon (A), Webcam (W), DSLR (D), and Caltech (C).

\begin{table*}[t]
  \centering
  \caption{\newmodel's average test accuracy is evaluated against multiple transformer-based approaches across diverse \noniid scenarios, showcasing its robustness and effectiveness relative to other advanced methods.
}
  \scalebox{0.68}{%
    \setlength\tabcolsep{5pt} 
    \begin{tabular}{lcccccccc}
      \toprule
      & & \multicolumn{2}{c}\textbf{RSNA} dataset & & & \multicolumn{2}{c}\textbf{Kermany} dataset\\
      \cmidrule(lr){2-5} \cmidrule(lr){6-9} 
      \# distribution & {Pathological} & {Pathological} & {Beta} & {Beta} & {Pathological} & {Pathological} & {Beta} & {Beta} \\
      \cmidrule(lr){2-5} \cmidrule(lr){6-9} 
      \# no. of clients & {100} & {200} & {100} & {200} & {100} & {200} & {100} & {200} \\
      \midrule
      \localt & 84.55$\pm0.15$ & 82.21$\pm0.08$ & 69.94$\pm0.13$ & 66.68$\pm0.13$ & 55.91$\pm0.17$ & 49.25$\pm0.11$ & 27.87$\pm0.12$ & 23.34$\pm0.10$ \\
      \fedavgt & 50.42$\pm4.22$ & 46.28$\pm4.23$ & 61.85$\pm1.50$ & 59.23$\pm1.93$ & 34.02$\pm0.88$ & 30.20$\pm0.95$ & 38.64$\pm0.22$ & 34.89$\pm0.45$ \\
      \fedpert & 89.86$\pm0.89$ & 89.01$\pm0.12$ & 79.41$\pm0.16$ & 77.70$\pm0.14$ & 67.23$\pm0.32$ & 61.72$\pm0.16$ & 37.19$\pm0.18$ & 29.58$\pm0.14$ \\
      \fedtpt & 79.75$\pm0.22$ & 75.46$\pm0.11$ & 77.25$\pm0.69$ & 71.13$\pm0.84$ & 48.61$\pm0.45$ & 46.05$\pm0.47$ & 36.63$\pm0.98$ & 25.13$\pm0.35$ \\
      \fedhnt & 82.26$\pm0.61$ & 77.57$\pm0.52$ & 71.45$\pm0.87$ & 68.13$\pm0.67$ & 53.08$\pm0.72$ & 39.94$\pm0.91$ & 33.25$\pm0.77$ & 29.14$\pm0.98$ \\
      \midrule
      \rowcolor[gray]{0.92}
      Vanilla-T & 91.83$\pm0.27$ & 91.28$\pm0.12$ & 89.23$\pm0.78$ & 87.77$\pm0.37$ & 88.67$\pm0.54$ & 88.23$\pm0.11$ & 87.74$\pm0.12$ & 87.26$\pm0.85$ \\
      \rowcolor[gray]{0.85}
      \textbf{\newmodel} & \textbf{92.67$\pm$0.74} & \textbf{91.34$\pm$0.86} & \textbf{88.49$\pm$0.38} & \textbf{88.16$\pm$0.33} & \textbf{89.80$\pm$0.23} & \textbf{87.73$\pm$0.74} & \textbf{87.34$\pm$0.92} & \textbf{86.98$\pm$0.64} \\
      \bottomrule
    \end{tabular}%
  }
  \label{tab_transformer}
\end{table*}

\begin{table*}[t]
  \centering
  \caption{Evaluation of \newmodel and Benchmark Methods on Image Datasets Under Various \noniid Settings}
  \scalebox{0.6}{%
    \setlength\tabcolsep{5pt} 
    \begin{tabular}{lcccccccc}
      \toprule
      & & \multicolumn{2}{c}{RSNA dataset} & & & \multicolumn{2}{c}{Kermany dataset} \\
      \cmidrule(lr){2-5} \cmidrule(lr){6-9} 
      \# distribution & {Pathological} & {Pathological} & {Beta} & {Beta} & {Pathological} & {Pathological} & {Beta} & {Beta} \\
      \cmidrule(lr){2-5} \cmidrule(lr){6-9} 
      \# no. of clients & {100} & {200} & {100} & {200} & {100} & {200} & {100} & {200} \\
      \midrule
      \fedavg~\cite{mcmahan2017communication}& 37.46$\pm0.81$ & 39.86$\pm0.85$ & 52.80$\pm0.95$ & 48.80$\pm0.18$ & 31.88$\pm0.08$ & 28.80$\pm0.26$ & 48.82$\pm0.31$ & 38.72$\pm0.89$ \\
      \fedper~\cite{arivazhagan2019federated}& 82.46$\pm0.91$ & 81.56$\pm0.17$ & 78.71$\pm0.08$ & 75.61$\pm0.43$ & 59.08$\pm0.77$ & 60.34$\pm0.11$ & 38.08$\pm0.44$ & 28.41$\pm0.78$ \\
      \fedme~\cite{t2020personalized}& 83.81$\pm0.48$ & 85.91$\pm0.07$ & 75.14$\pm0.38$ & 78.22$\pm0.92$ & 62.72$\pm0.08$ & 61.84$\pm0.70$ & 38.94$\pm0.09$ & 31.55$\pm0.16$ \\
      \fedprox~\cite{li2020federated}& 39.87$\pm0.64$ & 44.50$\pm0.90$ & 56.73$\pm0.25$ & 49.06$\pm0.24$ & 34.71$\pm0.46$ & 30.84$\pm0.34$ & 51.66$\pm0.83$ & 39.28$\pm0.77$ \\
      \fedbn~\cite{li2021fedbn} & 85.41$\pm0.84$ & 87.65$\pm0.74$ & 81.65$\pm0.31$ & 79.98$\pm0.04$ & 66.41$\pm0.30$ & 63.25$\pm0.73$ & 48.36$\pm0.40$ & 48.89$\pm0.36$ \\
      \fedhn~\cite{ha2016hypernetworks}& 88.12$\pm0.32$ & 89.51$\pm0.87$ & 83.13$\pm0.30$ & 83.06$\pm0.54$ & 55.91$\pm0.17$ & 49.25$\pm0.11$ & 27.87$\pm0.12$ & 23.34$\pm0.10$ \\
      \fedtp~\cite{li2023fedtp}& 87.85$\pm0.44$ & 88.36$\pm0.50$ & 85.63$\pm0.30$ & 80.65$\pm0.41$ & 68.99$\pm0.25$ & 69.25$\pm0.74$ & 51.78$\pm0.05$ & 48.85$\pm0.11$ \\
      \fedgp~\cite{achituve2021personalized}& 89.77$\pm0.21$ & 89.01$\pm0.78$ & 85.44$\pm0.39$ & 81.44$\pm0.41$ & 77.43$\pm0.37$ & 78.36$\pm0.26$ & 63.85$\pm0.22$ & 57.41$\pm0.91$ \\
      \fedrod~\cite{chen2021bridging}& 89.54$\pm0.75$ & 90.27$\pm0.07$ & 84.25$\pm0.36$ & 85.16$\pm0.98$ & 81.78$\pm0.17$ & 83.74$\pm0.98$ & 73.87$\pm0.75$ & 66.23$\pm0.75$ \\
      \midrule
      \rowcolor[gray]{0.85}
      \textbf{\newmodel (Ours)} & \textbf{92.67$\pm$0.74} & \textbf{91.34$\pm$0.86} & \textbf{88.49$\pm$0.38} & \textbf{88.16$\pm$0.33} & \textbf{89.80$\pm$0.23} & \textbf{87.73$\pm$0.74} & \textbf{87.34$\pm$0.92} & \textbf{86.98$\pm$0.64} \\
      \bottomrule
    \end{tabular}%
  }
  \label{tab_rnsa}
\end{table*}

\subsubsection{\newmodel Setup}
Following the experimental setup described in \fedhn, we performed experiments using \newmodel and benchmark methods with 100 and 200 clients. For the \textbf{Kermany} dataset, we considered 5\% participation, while for the \textbf{RSNA} dataset, we considered 10\% participation. The image classification task involved training each algorithm for 2000 communication rounds. \fedhn was trained for 4000 global communication rounds to ensure equivalent communication costs. For the detection task, the methods were trained for 300 communication rounds.

Both tasks used the \texttt{SGD} optimizer with a learning rate ($\text{lr}$) of 0.01 and a batch size ($B$) of 32. In \newmodel, the Learnable generators were optimized similarly with $\beta = 0.01$. Experiments were conducted on a cluster with an NVIDIA Tesla V100 GPU, simulating both the server and clients.

\subsection{Implementation Details}
Following the experimental setup of \fedhn~\cite{ha2016hypernetworks}, we implemented \fedtp~\cite{li2023fedtp} along with benchmark methods under standardized conditions. For the \textbf{CIFAR-10} dataset, we used 50 clients with a participation rate of 10\%, while for \textbf{CIFAR-100}, we used 100 clients with a participation rate of 5\%. The image classification models were trained for 1500 communication rounds, whereas \fedhn was trained for 5000 rounds to maintain equivalent communication costs.  
For the next-character prediction task, all methods were trained for 300 communication rounds. We employed the \texttt{SGD} optimizer across all tasks, using a learning rate of 0.01, a batch size of 64, and 5 local epochs per round. In \fedtp and \fedhn, hypernetworks were optimized with \texttt{SGD} using $\beta = 0.01$.  
All experiments were conducted on a computing cluster equipped with an RTX 2080 Ti GPU, with implementations carried out in PyTorch.

\subsubsection{Performance Analysis on Pneumonia Datasets}
We conducted a comprehensive performance comparison between \newmodel and several well-known federated learning methods, designed initially based on \cnn backbones. Table \ref{tab_transformer} displays the average test accuracy of these algorithms, highlighting \newmodel's remarkable performance superiority over each of them. This result strongly supports the assertions made in our Introduction section: \ref{1}) The \fedavg algorithm could impede the distinctive client representations within transformer models, as evidenced by \localt outperforming \fedavgt; and \ref{2}) \newmodel's learned customized focal modulation effectively addresses data heterogeneity. As depicted in Table \ref{tab_transformer}, \newmodel consistently outperforms Vanilla customized-T across all settings, thereby confirming that the ``learn-to-tailor" approach leverages the strengths of focal modulation in transformer models. 
To deepen our understanding, we evaluated test accuracy along with the curve depicting global communication rounds in \newmodel as shown in Table~\ref{tab_rnsa}. The analysis reveals that \newmodel demonstrates a smooth curve and achieves higher accuracy compared to alternative approaches.

\begin{table*}[htbp]
\caption{Comparison of Results: \newmodel and Benchmark Methods on Image Datasets with Varied \noniid Settings.}
  \resizebox{\textwidth}{!}{%
 
  \begin{tabular}{lcccccccc}
    \toprule
    & & \multicolumn{2}{c}{CIFAR 10} & & & \multicolumn{2}{c}{CIFAR 100}  \\
    \midrule  
    settings &
    \multicolumn{2}{c}{Pathological} &
    \multicolumn{2}{c}{Dirichlet} &
    \multicolumn{2}{c}{Pathological} & \multicolumn{2}{c}{Dirichlet} \\
    Client & 50 & 100 & 50 & 100 & 50 & 100 & 50 & 100\\
    \midrule
    \fedavg \cite{mcmahan2017communication} & 47.79$\pm$4.48 & 44.12$\pm$3.10 & 56.59$\pm$0.91 & 57.52$\pm$1.01 & 15.71$\pm$0.35 & 14.59$\pm$0.40 & 18.16$\pm$0.58 & 20.34$\pm$1.34 \\
    
   \fedprox \cite{li2020federated} & 50.81$\pm$2.94 & 57.38$\pm$1.08 & 58.51$\pm$0.65 & 56.46$\pm$0.66 & 19.39$\pm$0.63 & 21.32$\pm$0.71 & 19.18$\pm$0.30 & 19.40$\pm$1.76 \\
   
    \fedper \cite{arivazhagan2019federated} & 83.39$\pm$0.47 & 80.99$\pm$0.71 & 77.99$\pm$0.02 & 74.21$\pm$0.07 & 48.32$\pm$1.46 & 42.08$\pm$0.18 & 22.60$\pm$0.59 & 20.06$\pm$0.26 \\
    
    \fedme \cite{t2020personalized} & 86.09$\pm$0.32 & 85.23$\pm$0.58 & 76.29$\pm$0.44 & 74.83$\pm$0.28 & 49.09$\pm$1.10 & 45.57$\pm$1.02 & 31.60$\pm$0.46 & 25.43$\pm$0.52 \\
    
   \fedbn \cite{li2021fedbn} & 87.45$\pm$0.95 & \textcolor{gray}{86.71$\pm$0.56} & 74.63$\pm$0.60 & 75.41$\pm$0.37 & 50.01$\pm$0.59 & 48.37$\pm$0.56 & 28.81$\pm$0.50 & 28.70$\pm$0.46 \\
   
    \fedhn \cite{ha2016hypernetworks} & \textcolor{gray}{88.38$\pm$0.29} & 87.97$\pm$0.70 & 71.79$\pm$0.57 & 68.36$\pm$0.86 & \textcolor{gray}{59.48$\pm$0.67} & \textcolor{gray}{53.24$\pm$0.31} & 34.05$\pm$0.41 & 29.87$\pm$0.69 \\
    
    \fedgp \cite{achituve2021personalized} & 89.20$\pm$0.30 & 88.80$\pm$0.20 & 80.44$\pm$0.83 & \textcolor{gray}{78.29$\pm$1.13} & 63.30$\pm$0.10 & 48.98$\pm$0.01 & \textcolor{gray}{34.28$\pm$0.34} & \textcolor{gray}{28.29$\pm$1.53} \\
    
   \fedtp \cite{li2023fedtp} & \textcolor{gray}{90.31$\pm$0.26} & 88.39$\pm$0.14 & \textcolor{gray}{81.24$\pm$2.17} & 80.27$\pm$0.28 & 68.05$\pm$0.24 & 63.76$\pm$0.39 & 46.35$\pm$0.29 & 43.74$\pm$0.39 \\
   \fedrod \cite{chen2021bridging} & 89.87$\pm$0.03 & \textcolor{gray}{89.05$\pm$0.04} & 75.01$\pm$0.09 & 73.99$\pm$0.09 & 63.30$\pm$0.10 & 61.30$\pm$0.20 & 27.04$\pm$0.73 & 28.29$\pm$1.53 \\
   \midrule
   \rowcolor[gray]{0.85}
    \multicolumn{1}{l}{\textbf{\newmodel (Ours)}} &
\multicolumn{1}{r}{\textbf{93.47$\boldsymbol{\pm}$0.75}} &
\multicolumn{1}{r}{\textbf{91.85$\boldsymbol{\pm}$0.39}} &
\multicolumn{1}{r}{\textbf{82.89$\boldsymbol{\pm}$0.75}} &
\multicolumn{1}{r}{\textbf{79.75$\boldsymbol{\pm}$0.15}} &
\multicolumn{1}{r}{\textbf{71.96$\boldsymbol{\pm}$0.54}} &
\multicolumn{1}{r}{\textbf{68.11$\boldsymbol{\pm}$0.39}} &
\multicolumn{1}{r}{\textbf{51.75$\boldsymbol{\pm}$0.12}} &
\multicolumn{1}{r}{\textbf{44.33$\boldsymbol{\pm}$0.74}} \\
   
    \bottomrule
  \end{tabular}%
   }

\label{tab_non_iid}
\end{table*}

\begin{figure*}[t]
    \centering
    \includegraphics[width=\linewidth]{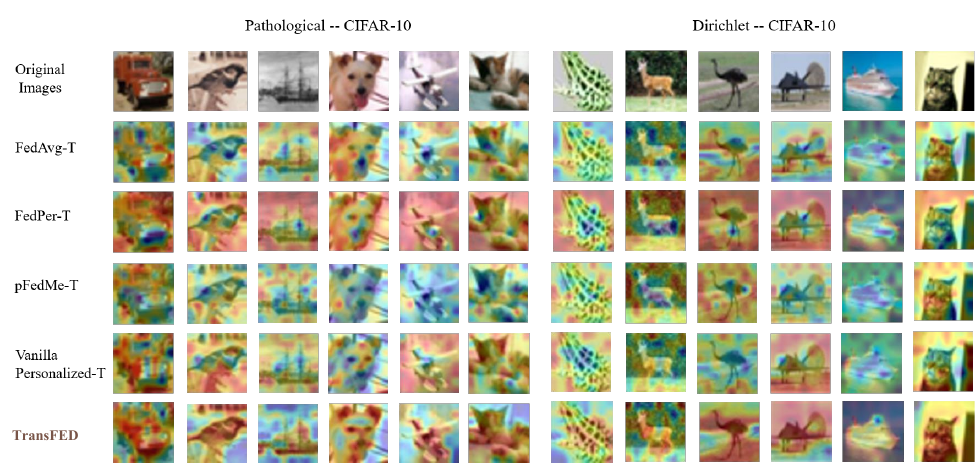}
    \caption{Saliency maps of \newmodel and other Transformer-based variants evaluated on \textbf{CIFAR-10} with 50 clients.}
    \label{fig:saliency}
\end{figure*}

\subsubsection{CIFAR 10 and CIFAR 100}
In our comprehensive comparative analysis of the experimental outcomes, as detailed in Table \ref{tab_non_iid}, our novel \newmodel
model emerges as a standout performer in direct contrast to the state-of-the-art benchmark methods. Across a diverse
range of \noniid settings utilizing the \textbf{CIFAR-10} and \textbf{CIFAR-100} datasets, \newmodel consistently demonstrates superior
accuracy rates. This distinctive performance advantage is particularly evident when considering various client populations,
encompassing both 50 and 100 clients. The noteworthy enhancement \newmodel brings over \fedtp significantly emphasizes
the effectiveness of our approach in managing personalized and task-specific learning objectives within the broader context
of federated learning. These findings strongly validate \newmodel’s ability to excel in complex scenarios characterized by
heterogeneous and personalized data distributions, effectively positioning it as a highly promising solution for effectively
addressing the inherent challenges within federated learning paradigms.  Figure~\ref{fig:saliency} presents a visual 
 \begin{table*}[t]
    \centering
    \caption{Classification accuracy (\%) on the Office-Caltech dataset~\cite{griffin2007caltech} using various federated domain adaptation (DA) methods, with baseline results reproduced from~\cite{peng2019federated}.
}
    \resizebox{0.8\textwidth}{!}{
    \begin{tabular}{lcccccc}
        \toprule
        Method & C,D,W$\rightarrow$A & A,D,W$\rightarrow$C & A,C,W$\rightarrow$D & A,C,D$\rightarrow$W & Avg \\
        \midrule
        \texttt{ResNet101}~\cite{he2016deep} & 81.9 & 87.9 & 85.7 & 86.9 & 85.6 \\
        \texttt{AdaBN}~\cite{li2016revisiting} & 82.2 & 88.2 & 85.9 & 87.4 & 85.7 \\
        \texttt{AutoDIAL}~\cite{maria2017autodial} & 83.3 & 87.7 & 85.6 & 87.1 & 85.9 \\
        \texttt{f-DAN1}~\cite{peng2019federated,long2015learning} & 82.7 & 88.1 & 86.5 & 86.5 & 85.9 \\
        \texttt{f-DANN2}~\cite{peng2019federated,ganin2015unsupervised} & 83.5 & 88.5 & 85.9 & 87.1 & 86.3 \\
        \texttt{FADA}~\cite{peng2019federated} & 84.2 & 88.7 & 87.1 & 88.1 & 87.1 \\
        \texttt{FedRF\_TCA}~\cite{feng2023robust} & 94.1 & 98.1 & 98.9 & 88.9 & 95.0 \\
        \midrule
        \rowcolor[gray]{0.85}
   \textbf{\newmodel (Ours)} & \textbf{95.6} & \textbf{98.3} & \textbf{99.1} & \textbf{93.4} & \textbf{96.9} \\
        \bottomrule
    \end{tabular}
    }
    \label{tab_fedsfda}
\end{table*}
 comparison of attention maps between \fedtp and other Transformer-based variants on CIFAR-10 and CIFAR-100. Both Vanilla Personalized-T and \newmodel exhibit client-specific self-attention, producing distinct and interpretable visualization maps. Notably, \newmodel consistently emphasizes critical regions of test objects more effectively than PFedME and \fedavg, demonstrating its ability to capture personalized attention patterns. These results further validate \newmodel’s effectiveness in adapting to client heterogeneity and enhancing model performance.

\subsubsection{Shakespeare Dataset}
We compared \newmodel with state-of-the-art federated learning methods using \texttt{CNN/LSTM} architectures. Table \ref{tab_shakespeare} summarizes their average test accuracy across image and language datasets. \newmodel consistently outperforms baselines, highlighting its effective personalized learning and robust parameter optimization. Its efficient parameterization reduces computational overhead and enhances adaptability to heterogeneous client data, achieving superior performance over the state-of-the-art.

\subsubsection{Federated Source-free domain adaptation (\fsfda)}
Table~\ref{tab_fedsfda} presents a comprehensive evaluation of \newmodel, showcasing its state-of-the-art (\sota) performance against benchmark federated domain adaptation (DA) methods. Additionally, we assess the robustness of \newmodel in the presence of message and client dropouts caused by unstable network conditions.  
Comparing settings (II) and (III) to setting (I) in Table~\ref{tab_fedsfda}, we observe that restricting communication to a randomly selected subset \( S_t \) of source clients and asynchronously aggregating the classifier does not degrade \newmodel’s performance. Furthermore, \newmodel achieves optimal results on the \textbf{Office-Caltech} dataset, further demonstrating its effectiveness and resilience in diverse federated learning scenarios.

\begin{wraptable}{r}{0.58\linewidth} 
  \vspace{-\baselineskip}
  \centering
  \caption{Comparative Analysis of Average Test Accuracy for Various Methods on the \textbf{Shakespeare} Language Dataset.}
  \label{tab_shakespeare}
  \setlength{\tabcolsep}{4pt}         
  \renewcommand{\arraystretch}{1.05}  
  \small                               
  \resizebox{\linewidth}{!}{
  \begin{tabular}{l|c|c}
    \toprule
    \rowcolor[gray]{0.92}
    Method & Venue & Test Accuracy \\
    \midrule
    \fedavg~\cite{mcmahan2017communication} & PMLR'17 & $56.86 \pm 0.86$ \\
    \fedprox~\cite{li2020federated}         & MLS'20  & $56.68 \pm 1.09$ \\
    \fedper~\cite{arivazhagan2019federated} & Arxiv'19& $68.71 \pm 0.65$ \\
    \fedme~\cite{t2020personalized}         & NeurIPS'20& $63.14 \pm 1.12$ \\
    \texttt{MIME SGD}~\cite{karimireddy2020mime}  & ICML'21 & $56.26 \pm 0.25$ \\
    \texttt{MIME SGDm}~\cite{karimireddy2020mime} & ICML'21 & $54.00 \pm 1.28$ \\
    \texttt{MIMELite}~\cite{karimireddy2020mime}  & ICML'21 & $52.48 \pm 1.12$ \\
    \texttt{FedCM}~\cite{xu2021fedcm}       & IJCNN'21  & $38.90 \pm 1.12$ \\
    \texttt{SCAFFOLD}~\cite{karimireddy2020scaffold} & PMLR'20 & $56.68 \pm 1.66$ \\
    \texttt{FedDyn}~\cite{acar2021federated} & ICLR'21   & $54.54 \pm 0.64$ \\
    \texttt{AdaBest}~\cite{varno2022adabest} & ECCV'22   & $58.12 \pm 0.18$ \\
    \texttt{FedACG}~\cite{kim2024communication} & CVPR'24 & $56.79 \pm 1.25$ \\
    \midrule
    \rowcolor[gray]{0.85}
    \textbf{\newmodel (Ours)} & & \textbf{87.57} $\pm$ \textbf{0.47} \\
    \bottomrule
  \end{tabular}
  }
  \vspace{-0.5\baselineskip}
\end{wraptable}

\subsubsection{Analysis of Different Adapted Parts}
This study examined the effects of personalizing various components of the transformer model. Specifically, we focused on four components: (1) the focal-modulation layers (our proposed method), (2) the Multi-layer perceptron layers, (3) the normalization layers, and (4) the encoder. To maintain a fair comparison, we employed the identical Learnable generator to generate the parameters associated with these individual components while ensuring consistency in the \focalnet structures as described. The results of this experiment are presented in Table \ref{tab_components}. It is evident from the table that personalizing the focal-modulation layers yields the best performance compared to personalizing other components. 
\begin{table}[t]
\centering
\caption{The Test Accuracy Averages Across 100 Clients for Federated Transfer Learning with Pre-trained Models (\model) and Its Variants.}
\scalebox{0.80}{
\begin{tabular}{lcccc}
\toprule
\multirow{2}{*}{setting} & \multicolumn{2}{c}{RSNA} & \multicolumn{2}{c}{Kermany} \\
\cmidrule(lr){2-3} \cmidrule(lr){4-5}
& {Pathological} & {Beta} & {Pathological} & {Beta} \\
\midrule
\rowcolor[gray]{0.85}
\model (ours) & 92.67 $\pm$ 0.74 & 88.49 $\pm$ 0.38 & 89.80 $\pm$ 0.23 & 87.34 $\pm$ 0.92 \\
\rowcolor[gray]{0.95}
\model+\knn & 82.34 $\pm$ 0.43 & 83.65 $\pm$ 0.52 & 83.79 $\pm$ 0.24 & 83.95 $\pm$ 0.37 \\
\rowcolor[gray]{0.95}
\model+\fedper& 88.45 $\pm$ 0.14 & 86.36 $\pm$ 0.17 & 87.76 $\pm$ 0.14 & 85.97 $\pm$ 0.16 \\
\rowcolor[gray]{0.95}
\model+\fedrod & 89.56 $\pm$ 0.45 & 86.55 $\pm$ 0.27 & 86.23 $\pm$ 0.37 & 87.22 $\pm$ 0.39 \\
\bottomrule
\end{tabular}}
\label{tab_varients}
\end{table}

Furthermore, Table \ref{tab_components} illustrates that customizing the normalization layers yields superior performance compared to tailoring the \mlp layers and the absolute encoder.
\subsubsection{Generalization to Novel Clients}

We thoroughly assessed our method's capacity for generalization, contrasting it with \fedme, \fedhn, \fedrod, and a customized-T Vanilla approach on the \textbf{Kermany} and \textbf{RSNA} datasets under the Beta configuration. To simulate a realistic scenario, 30\% of the clients were randomly selected as novel clients whose data had not been seen during the training phase. \fedper fine-tuned the customized parameters in the last classification layer, while \fedme learned all parameters to obtain customized models for each client. In the case of \fedhn and \newmodel, the customized parameters had the option to choose between clients embedding vectors with a dimension of 32 and the entire Learnable generator. As presented in Table \ref{tab_generalization1}, the outcomes suggest that \newmodel (Adaptive Learnable generator) exhibits improved resilience and adeptly adjusts to new clients in a few epochs.

\begin{table}[t]
\centering
\caption{Generalization Performance Comparison on \textbf{RSNA} dataset.}
\label{tab_generalization1}
\scalebox{0.79}{%
\begin{tabular}{lccc}
\toprule
\rowcolor[gray]{0.88}
\textbf{Method} & \textbf{Personalization} & \textbf{Client Acc (\%)} & \textbf{Epochs} \\
\midrule
\fedme~\cite{t2020personalized} & All Parameters & 78.3 & 8 \\
\fedtp~\cite{li2023fedtp} & Self Attention Layers & 81.2 & 4 \\
\fedhn (Embedding)~\cite{ha2016hypernetworks} & Clientwise Embedding & 79.5 & 6 \\
\fedhn (Hypernetwork)~\cite{ha2016hypernetworks} & Whole Hypernetwork & 80.2 & 5 \\
\fedrod~\cite{chen2021bridging} & Last Classification Layer & 77.8 & 10 \\
Vanilla Personalized-T & SA Projection Matrices & 76.7 & 12 \\
\rowcolor[gray]{0.95}
\newmodel (Learnable Generator) & Focal Modulation Layers & \textbf{82.6} & \textbf{3} \\
\bottomrule
\end{tabular}}
\end{table}

\subsubsection{Analysis of Adaptive Learnable generators}
To thoroughly examine the impact of adaptive Learnable generators, we conducted a comparative analysis between \newmodel and Vanilla customized-T. The latter method restores the projection parameters $P_i$ for each client locally, without the utilization of Learnable generators. As depicted in Table \ref{tab_transformer}, \newmodel exhibits a significant advantage over Vanilla customized-T, highlighting the crucial role of Learnable generators in \newmodel. We also observed that even when Learnable generators solely generate the parameters of the focal-modulation layer, they effectively encode client-specific information into client embeddings $z_i$. The Learnable generators can project client embeddings $z_i$ onto a manifold defined by the parameters $\phi$ of the Learnable generator.

\begin{wrapfigure}{r}{0.48\linewidth}
  \centering
  \vspace{-\baselineskip} 
  \includegraphics[width=\linewidth]{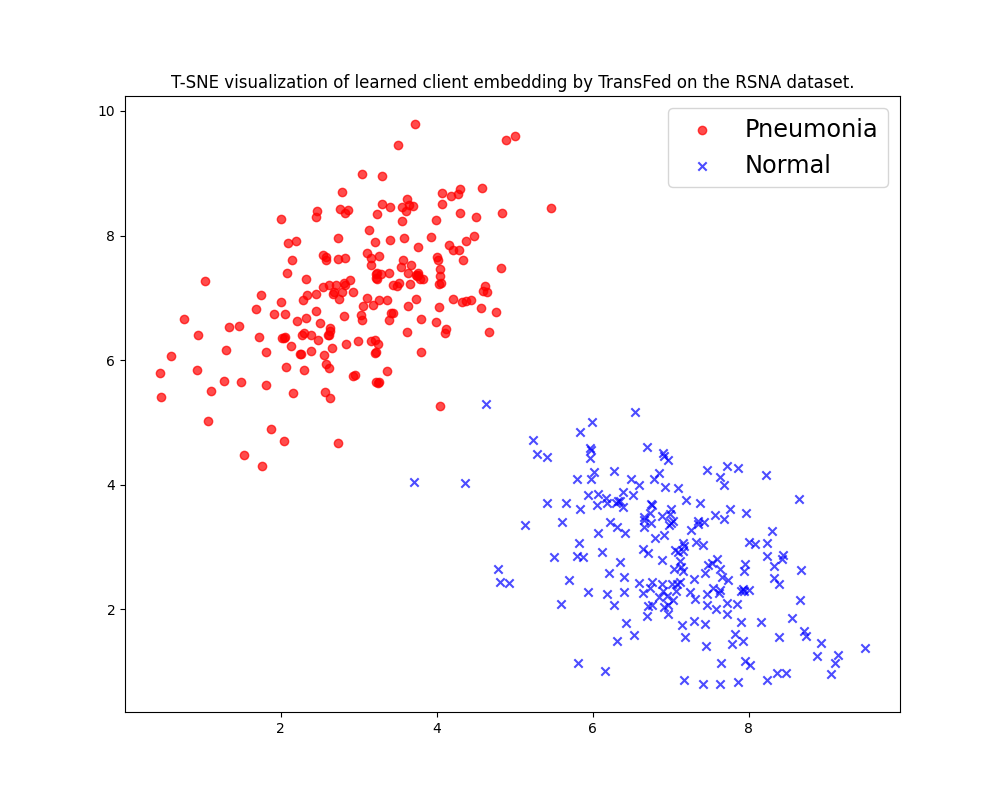}
  \caption{Visualization of client embeddings learned by \newmodel using t-SNE on the \textbf{RSNA} dataset.}
  \label{fig:tsne}
  \vspace{-0.9\baselineskip} 
\end{wrapfigure}

To delve deeper into the acquired client embeddings, we utilized the \texttt{t-SNE} algorithm for their projection onto a 2-D plane, as illustrated in Figure \ref{fig:tsne}. Specifically, we distributed each coarse class among five clients, ensuring that the corresponding fine classes were uniformly allocated among these selected clients. We also trained \newmodel and visualized the client embeddings after training. Learned individual embeddings of clients who share common coarse labels tend to cluster together, while those with dissimilar coarse labels are mapped farther apart. This outcome provides compelling evidence supporting our claim that Learnable generators are highly effective for encoding customized information into client embeddings $z_i$.

\subsection{Ablation Study}
\subsubsection{Extension}
In this section, we investigate the compatibility of \newmodel with existing methods that utilize personalized classifier heads such as \fedper~\cite{arivazhagan2019federated}, \fedrod~\cite{chen2021bridging}, and methods employing local memory like \knnper~\cite{marfoq2022personalized}. These modules were integrated into \newmodel and evaluated on the \textbf{CIFAR-10/CIFAR-100} datasets. 
For \newmodel+\fedper, each client retains its classification head locally within the \newmodel framework. \newmodel+\fedrod combines the output of the personalized classification head and the global classification head to compute prediction logits. \newmodel+\knnper establishes and maintains a local repository akin to \knnper, utilizing the \texttt{FAISS} library~\cite{johnson2019billion}. 
Table \ref{tab_varients} presents the results of these experiments.

\subsubsection{Effect of Heterogeneity in Label Distribution:}
Data heterogeneity, particularly in label distribution, presents a significant challenge in personalized federated learning. To analyze its impact, we conducted experiments on the \textbf{RSNA} and \textbf{Kermany} datasets while varying the Beta distribution parameter $\alpha$. Our results demonstrate that \newmodel consistently outperforms state-of-the-art (\sota) methods. In previous experiments, we examined label distribution heterogeneity by sampling class distributions from a Dirichlet distribution with $\alpha = 0.3$. To extend this analysis, we now consider a broader range of cases with $\alpha = \{0.1, 0.3, 0.5, 0.7, 0.9\}$ on the \textbf{CIFAR-10} and \textbf{CIFAR-100} datasets, where smaller $\alpha$ values indicate higher data heterogeneity.  
Using \fedavgt as a baseline, we compared \newmodel against personalized federated learning methods, including \fedpert, \fedhn, and \fedbn. As shown in Table~\ref{tab_heteregenous}, \fedavgt exhibits decreasing performance as data heterogeneity increases, while methods incorporating personalized modules demonstrate improved robustness. Among these approaches, \newmodel consistently achieves the best performance across all levels of data heterogeneity, highlighting its ability to effectively mitigate challenges posed by non-IID label distributions.  
As $\alpha$ increases, certain personalized federated learning methods struggle to fully exploit client heterogeneity, occasionally performing worse than \fedavgt in accuracy. However, \newmodel maintains strong and stable performance, underscoring its capability to leverage client-specific data distributions effectively, even when heterogeneity is less pronounced.
\begin{table}[t]
\centering
\caption{Average test accuracy of models with varying customized components.}
\scalebox{0.79}{
\begin{tabular}{lcccc}
\toprule
\multirow{2}{*}{Customized Part} & \multicolumn{2}{c}{RSNA} & \multicolumn{2}{c}{Kermany} \\
\cmidrule(lr){2-3} \cmidrule(lr){4-5}
& {Pathological} & {Beta} & {Pathological} & {Beta} \\
\midrule
\rowcolor[gray]{0.88}
Focal Modulation & \textbf{92.67 $\pm$ 0.74} & \textbf{88.49 $\pm$ 0.38} & \textbf{89.80 $\pm$ 0.23} & \textbf{87.34 $\pm$ 0.92} \\
\rowcolor[gray]{0.92}
MLP Layers & 88.45 $\pm$ 0.14 & 86.36 $\pm$ 0.17 & 87.76 $\pm$ 0.14 & 85.97 $\pm$ 0.16 \\
\rowcolor[gray]{0.89}
Normalization Layers & 89.56 $\pm$ 0.45 & 86.55 $\pm$ 0.27 & 86.23 $\pm$ 0.37 & 87.22 $\pm$ 0.39 \\
\rowcolor[gray]{0.98}
Encoder & 82.34 $\pm$ 0.43 & 83.65 $\pm$ 0.52 & 83.79 $\pm$ 0.24 & 83.95 $\pm$ 0.37 \\
\bottomrule
\end{tabular}}
\label{tab_components}
\end{table}

\begin{table*}[t]
    \centering
    \caption{Results of \model and Benchmark Methods on Image Datasets Over 100 Clients with Different $\alpha$ of Dirichlet Setting. Red color denotes second highest.}
    \label{tab_heteregenous}
    \resizebox{\textwidth}{!}{%
    \begin{tabular}{l*{6}{c}}
        \toprule
        & \multicolumn{3}{c}{RSNA Test Accuracy (\%)} & \multicolumn{3}{c}{Kermany Test Accuracy (\%)} \\
        \cmidrule(lr){2-4} \cmidrule(lr){5-7}
        $\alpha$ & 0.1  & 0.5  & 0.9 & 0.1 & 0.5  & 0.9 \\
        \midrule
        \rowcolor[gray]{0.95}
        \fedavgt~\cite{mcmahan2017communication} & $40.99 \pm 1.25$ & $39.23 \pm 0.63$ & $33.57 \pm 0.58$ & $45.29 \pm 1.54$ & $31.82 \pm 0.42$ & $29.14 \pm 0.81$  \\
        \fedpert~\cite{arivazhagan2019federated} & $77.45 \pm 0.14$ & $64.57 \pm 0.14$ & $62.44 \pm 0.22$ & $60.11 \pm 0.21$ & $51.13 \pm 0.14$ & $35.92 \pm 0.23$ \\
        \fedhn~\cite{ha2016hypernetworks} & $73.16 \pm 1.80$ & $70.47 \pm 0.38$ & {\color{red}$65.42 \pm 0.15$} & $51.61 \pm 0.04$ & $47.62 \pm 0.75$ & $36.45 \pm 0.50$ \\
        \fedtp~\cite{li2023fedtp} & $79.67 \pm 1.42$ & {\color{red}$71.36 \pm 0.86$} & $63.74 \pm 0.19$ & $58.19 \pm 0.81$ & $49.86 \pm 0.05$ & $37.37 \pm 0.95$  \\
        \fedbn~\cite{li2021fedbn} & {\color{red}$84.93 \pm 0.53$} & $62.41 \pm 0.37$ & $51.79 \pm 0.35$ & {\color{red}$65.57 \pm 0.12$} & {\color{red}$52.70 \pm 0.47$} & {\color{red}$44.10 \pm 0.84$} \\
        \rowcolor[gray]{0.85}
        \textbf{\model (Ours)} & $\mathbf{90.41 \pm 0.30}$ & $\mathbf{85.28 \pm 0.91}$ & $\mathbf{73.75 \pm 0.26}$ & $\mathbf{68.16 \pm 0.88}$ & $\mathbf{62.12 \pm 0.73}$ & $\mathbf{51.09 \pm 0.26}$ \\
        \bottomrule
    \end{tabular}%
    }
\end{table*}

Furthermore, we evaluated \newmodel’s resilience under increasing levels of Gaussian noise added to client data. Consistently, \newmodel outperformed competing methods, demonstrating superior robustness in handling client-specific noise.  
Additionally, we examined the effect of varying the number of focal modulation blocks in \newmodel. Our results, summarized in Table~\ref{tab_non_iid}, indicate that increasing the number of modulation blocks enhances the model’s ability to capture data heterogeneity and improves overall performance. Based on these findings, we set the default number of focal modulation blocks to eight in subsequent experiments.  
Finally, we investigated the impact of different client participation rates on model performance. Compared to \fedavgt, \newmodel exhibited greater stability across varying participation levels, reinforcing its robustness in federated learning settings with dynamic client availability.

\section{Conclusion and Future Work}
\label{5}
We presented \newmodel, a scalable and adaptable extension of \model that learns to personalize focal modulation layers via a centralized, learn-to-adapt hypernetwork. By conditioning on task-aware client embeddings, \newmodel tailors modulation dynamics to heterogeneous data distributions. We further provided enhanced theoretical guarantees with tighter adaptation/generalization bounds, and validated the approach across diverse modalities: including vision, time series, and multilingual text, demonstrating consistent improvements over \fedavg and strong personalized FL baselines, as well as over our original \textsc{TransFed}, in both source-free and cross-task federated settings. In addition, a low-rank hypernetwork conditioning variant reduces server–client communication overhead, enabling deployment in resource-constrained environments without sacrificing accuracy.

\noindent\textbf{Future Work.} Looking ahead, we plan to (i) study privacy-preserving embedding construction and auditing, (ii) integrate quantized/sparse hypernetwork updates for further efficiency, (iii) extend to continual and cross-task adaptation at larger scales, and (iv) tighten theory for non-convex transformers under realistic FL heterogeneity. We hope \newmodel and our codebase will facilitate more adaptive, scalable, and generalizable transformer-based federated systems.

\subsection{Data Availability}
We do not generate any datasets, because our work proceeds within a theoretical and mathematical approach. All data used for the experiments are publically available benchmark datasets. The code and model are released publicly.

\vskip 0.2in
\bibliography{sample}

\end{document}